\documentclass[a4paper]{article}
\usepackage{iwslt14,amssymb,amsmath,epsfig}
\usepackage{multirow,booktabs}
\usepackage{relsize}
\usepackage{paralist}

\def\reg{{\rm\ooalign{\hfil
     \raise.07ex\hbox{\scriptsize R}\hfil\crcr\mathhexbox20D}}}

\title{The USFD Spoken Language Translation System for IWSLT 2014}

\name{Raymond W. M. Ng, Mortaza Doulaty, Rama Doddipatla, Wilker Aziz, Kashif Shah,}{Oscar Saz, Madina Hasan, Ghada AlHarbi, Lucia Specia, Thomas Hain}

\address{Department of Computer Science, University of Sheffield, United Kingdom \\
{\small {\tt{ \{wm.ng,mortaza.doualty,r.doddipatla,w.aziz,kashif.shah,}}} \\
{\small {\tt{ o.saztorralba,m.hasan,GAlHarbi1,l.specia,t.hain\}@sheffield.ac.uk}}}
}
\begin{document}
\maketitle
\begin{abstract}
The University of Sheffield (USFD) participated in the International Workshop for Spoken Language Translation (IWSLT) in 2014. In this paper, we will introduce the USFD SLT system for IWSLT. Automatic speech recognition (ASR) is achieved by two multi-pass deep neural network systems with adaptation and rescoring techniques. Machine translation (MT) is achieved by a phrase-based system. The USFD primary system incorporates state-of-the-art ASR and MT techniques and gives a BLEU score of 23.45 and 14.75 on the English-to-French and English-to-German speech-to-text translation task with the IWSLT 2014 data. The USFD contrastive systems explore the integration of ASR and MT by using a quality estimation system to rescore the ASR outputs, optimising towards better translation. This gives a further 0.54 and 0.26 BLEU improvement respectively on the IWSLT 2012 and 2014 evaluation data.
\end{abstract}

\section{Introduction}

In this paper, the University of Sheffield (USFD) system for the International Workshop on Spoken Language Translation (IWSLT) 2014 is introduced. USFD participated in English-to-French and English-to-German SLT tasks. The ASR and MT systems made use of state-of-the-art technologies. On the ASR side, two deep neural network systems built on partially different data and different tandem configurations were used. On the MT side, phrase-based translation models were built. ASR and MT system integration attempts were made by using a translation quality estimation system. It considered the system scores from both ASR and MT, as well as features extracted from the ASR outputs in source language. The ASR hypotheses were then rescored based on the predicted translation quality. This gives performance improvements in terms of BLEU score increase.

In the following, the data used for system training is introduced in \S\ref{sec:data}. \S\ref{sec:ASR} and \S\ref{sec:MT} give the details of the ASR and MT systems. The decoding algorithm and system results are given in \S\ref{sec:decoding}. Besides the primary submission, USFD also submitted contrastive systems which implement system integration. These systems used a quality estimation module and performed ASR $N$-best list rescoring based on predicted translation quality. This would be described in \S\ref{sec:sysintegration}.

\section{Data processing and selection}\label{sec:data}

The ASR and MT systems were primarily trained on TED lecture data \cite{TED}. For ASR, TED and the additional data form two data subsets, on which two systems were trained. For MT, out-of-domain data after data selection were incorporated in the training of translation models and target language models. 

\subsection{ASR acoustic modelling}\label{ssec:data_am}

Two data sets were used for ASR system training. For the ease of discussion they are hereinafter referred to as ASR$_1$ and ASR$_2$. The composition of the two data sets is shown in Table \ref{table:data_am}. 

\begin{table}[h]
 \caption{Data for acoustic model training\label{table:data_am}}
 {\centering \centerline {\resizebox{0.88\linewidth}{!}{
 \begin{tabular}{lrllr}
 \hline
  \multicolumn{2}{c}{ASR$_1$}      &              & \multicolumn{2}{c}{ASR$_2$} \\
   Data  & Hours         & \hspace{1mm} & Data       & Hours\\
 \hline
 TED              & 132   &        & TED                  & 112 \\
 LLC              & 106   &        & AMI$+$AMIDA$+$ICSI\hspace{-4mm} & 165 \\
 ECRN             & 60    &        & ECRN                  & 60 \\
 \hline
 \end{tabular}
   }}}
\end{table}

TED serves as a common data set in both ASR$_1$ and ASR$_2$. Their segmentations in ASR$_1$ and ASR$_2$ differ slightly and this is explained later. The two data sets are augmented by e-corner lecture data (ECRN) with a duration of $60$ hours \cite{madina-interspeech2014}. ASR$_1$ also contains $106$ hours of LLC lecture data. In ASR$_2$, $165$ hours of meeting data from the AMI, AMIDA and ICSI corpora are added so the trained model will reflect also generic domains other than lectures \cite{AMIDA09,Doddipatla2014IS}.

The TED portions in both ASR$_1$ and ASR$_2$ originate from $734$ TED talks published before 31 Dec 2010. Each talk has a duration of around $15$ minutes. Human annotations in the form of subtitles  
are also available, giving rough segmentation with segment duration from $3$ to $5$ seconds and time accuracy to the nearest second. 

Exact segmentations and transcriptions of TED were derived in different ways in ASR$_1$ and ASR$_2$. In ASR$_1$, all segments from the same talk were merged and the speech was forced aligned, resegmented before another forced alignment run determined the final training set. This gave a total of $132$ hours of speech for AM training. In ASR$_2$, forced alignment was performed on the rough segmentation, after which contagious segments were merged when there was tight silence at the segment boundaries. A further run of forced alignment determined the final training set. This gave a total of $112$ hours of speech. 

To evaluate the performance of different segmentations, PLP-based state-tied triphone models with cepstral mean and variance normalisation were trained on these data and decoding was performed on the IWSLT 2010 evaluation data set. The WERs for the ASR$_1$ and ASR$_2$ settings are $25.7\%$ and $26.2\%$ respectively. When the models are trained directly on the roughly segmented data (no adjustment of segmentations), the total duration of training data is $109$ hours and the corresponding WER is $28.1\%$.

\subsection{Language models and MT}\label{ssec:data_text}

Textual data for the training of language models and translation models were obtained from the affiliated websites of the IWSLT and WMT evaluations \cite{WIT3_EAMT2012,WMT14WEB}. TED was considered as the in-domain training data and the full data set was used. Four out-of-domain (OOD) data sets from News commentary v9, Common Crawl, Gigaword and Europarl v7 were also used, after a data selection process.   

\begin{table}
 \caption{Amount of text data used in different training tasks in En$\rightarrow$Fr translation {\footnotesize ($^\#$Full data set was used for builing target LM)}\label{table:data_text}}
 {\centering \centerline {\resizebox{0.98\linewidth}{!}{
 \begin{tabular}{lrrrr}
  \hline
         &  \multicolumn{4}{c}{Number of words/million} \\
  Data   &  \hspace{-2mm}{\footnotesize Target LM$^\#$}\hspace{-1mm}  &  \hspace{-2mm}{\footnotesize Source LM}\hspace{-1mm}  & \hspace{-1mm}{\footnotesize Punct TM}\hspace{-2mm}  & \hspace{-2mm}{\footnotesize TM}\hspace{1mm} \\
  \hline
  TED                           &  3.17 & 3.17  & 3.17  & 3.17  \\
  News Commentary\hspace{-5mm}  &  4.0  & 0.9   & 0.2   & 0.7   \\
  Common crawl                  & 70.7  & 36.1  & 3.6   & 10.8  \\
  Gigaword                      & 575.7 & 271.2 & 26.3  & 14.9  \\
  Europarl                      & 50.3  & 10.8  & 4.3   & 1.9   \\
  \hline 
 \end{tabular}
  }}}
\end{table}

The OOD corpora were selected with the cross entropy difference criterion \cite{MooreLewis_2010ACL_dataselection}. Given a sentence $x_1^{I}=[x_1 \cdots x_I]$ with $I$ words, cross entropy values $H(x_1^I,ID)$ and $H(x_1^I,OOD)$ were computed using ${\cal G}_{\text{ID}}$, the ID language model (in this case, TED) and ${\cal G}_{\text{OOD}}$, the OOD language model (built on the corpus from which the sentence was taken). The cross entropy difference (CED) was given by,


\begin{equation}
 \text{CED}(x_1^{I}) = H(x_1^{I},{\cal G}_{\text{ID}}) - H(x_1^{I},{\cal G}_{\text{OOD}}) 
\end{equation}

\noindent Sentences were ranked by the CED values and 25\% of the sentences with the lowest CED values were selected from each corpus. Furthermore, CED values were calculated on sentence batches with increasing sizes. A line search was done to find the optimal batch giving the minimum $CED$ value. All data selection was done on the English text. For data selection to translation model training, the corresponding sentences in the target languages were extracted after selection was done on English sentences.

Table \ref{table:data_text} shows the amount of the full text data set, and the selected text data in different systems in the English$\rightarrow$French translation task. The full data set contains $703.9$M words. They were used for training the target language model in MT, which was a 5-gram interpolated LM with punctuation and out-of-vocabulary word modelling, modified Kneser-Ney smoothing and was in standard ARPA format. The source language model for ASR was built on the full TED data set and 25\% or 50\% of the OOD data, making up to $322.2$M words. A monolingual translation model was trained for punctuation insertion and case conversion. The training took the full TED data and 5-10\% of the OOD data, resulting in a total of $37.6$M words. The translation model was trained on the full TED data set and other optimally selected OOD data sets, where only around 5\% of the sentences were selected. The total number of words is $31.7$M. 

\begin{figure*}[t]
 \includegraphics[width=0.9\textwidth]{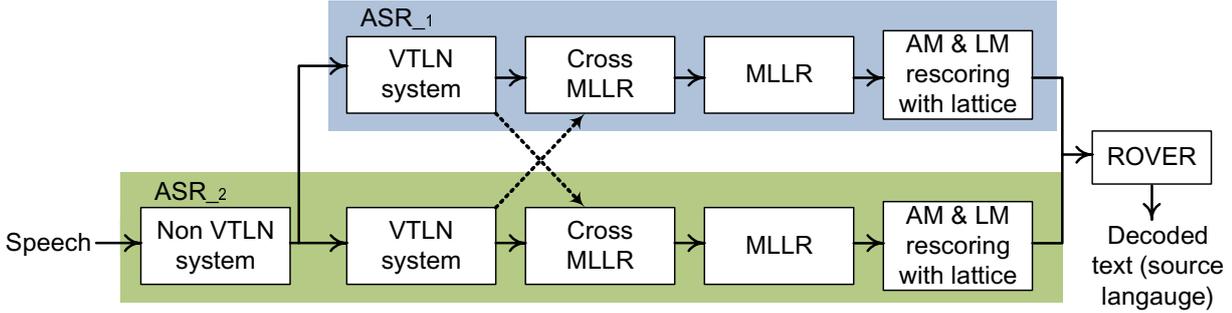}
 \caption{System diagram for multi-pass ASR decoding. \label{fig:ASRsysdiag}}
\end{figure*}

\section{Automatic speech recognition}\label{sec:ASR}

There are two DNN systems with tandem configurations in ASR \cite{Hermansky2000tandem}. Bottleneck (BN) features were derived from deep neural network (DNN)s \cite{Doddipatla2014IS}, and GMM-HMM systems were trained on these bottleneck features. The two tandem systems were trained on ASR$_1$ and ASR$_2$ data respectively (Table \ref{table:data_am}). Different portions of data were used in different stages of training. Let DNN$_1$ and DNN$_2$ denote the two DNN systems for ASR$_1$ and ASR$_2$. DNN$_1$ was trained on TED data only. DNN$_2$ was trained on TED and AMI$+$AMIDA$+$ICSI data only. The remaining data listed in Table \ref{table:data_am} were added to the training pool in the GMM-HMM training stage. 

DNN$_1$ has $4$ hidden layers, each having 1,745 hidden units. The BN layer is placed just before the output layer and has 26 units. The output layer has 4,320 units. DNN$_2$ has 5 hidden layers, with the first 3 layers having 1,745 units and the fourth hidden layer having 65 units. A BN layer is placed just before the output layer and has 39 units. The output layer has 5,691 units. 

Both the DNNs were trained using log filter-bank outputs and concatenating 31 adjacent frames, which were decorrelated using DCT to form a 368-dimensional feature vector. The filter-bank outputs were mean and variance normalised at the speaker level. 
Global mean and variance normalisation was performed on each dimension before feeding the input for training the DNN. The GMM-HMM systems trained using the BN features were different. 
The model for ASR$_1$ was trained on the concatenated features with the $26$-dimension BN features from DNN$_1$ and the $39$-dimension PLP features. The model for ASR$_2$ was trained on the $39$-dimension BN features from DNN$_2$. Both the GMM-HMM models were trained as tied-state triphone 
systems with the final models having $16$ mixture Gaussians per state.

All systems are vocal tract length normalised (VTLN). In the training stage, a PLP system was used to obtain the warp factors for each speaker. Then the filter-bank and PLP features were VTLN-warped, which were in turn used for DNN and GMM-HMM training in the tandem configuration. In the decoding stage,  a non-VTLN DNN and GMM-HMM tandem system trained on ASR$_2$ data replaced the PLP system for the derivation of warp factors. 

To improve the performance of the acoustic model, minimum phone error (MPE) training was performed using the lattices which were generated using a uni-gram language model \cite{Povey_2002ICASSP_MPE}.

Language models for ASR are all interpolated LMs built on the English text data described in Table \ref{table:data_text} and tuned on IWSLT 2010 dev and eval data. $2$-gram and $4$-gram ARPA language models were trained for lattice generation and expansion. The $4$-gram LM was pruned with a threshold $10^{-10}$ and a weighted-finite-state transducer (WFST) was constructed for fast decoding in the pre-final passes in the ASR systems. 

All ASR LMs were based on a word-list with a 60k word vocabulary extracted based on our standard English ASR inventory and the English part of the TED MT training data for IWSLT 2014 \cite{AMIDA09,WIT3_EAMT2012}. Pilot ASR experiments on the IWSLT 2011 and 2012 eval data show the drop of perplexity with the addition of Common crawl and Gigaword data. For these two corpora, the rate of data selected for LM building was set to $50\%$, while the rate for other OOD corpora was kept $25\%$. This made the total number of words $322.2$M as shown in Table \ref{table:data_text}. 

Pronunciation probabilities were incorporated in final stage decoding \cite{Hain_2005SpeCom_Pronprob}. These probabilities were extracted based on the Viterbi alignment of the phoneme level transcription of the ASR$_1$ training data. When a word allowed multiple pronunciations, the frequency of each pronunciation was calculated and stored. These frequencies were then applied to the words in the decoding dictionary for words that appeared in both training and decoding stages. Words with multiple pronunciations appearing only in the decoding stage were given equal probability.


\section{Machine translation}\label{sec:MT}

A phrase-based model using \textsc{Moses} \cite{Koehn+2007:moses} in a standard setting was employed. 
For phrase extraction all of the TED data ($3.17$ million words) was used.
Following previous findings \cite{UEDIN2013IWSLTMT}, 
data selection via a cross-entropy difference criterion (detailed in \S\ref{ssec:data_text}) was used to select the optimal batch of the OOD data, which amounts to about $5\%$ of the total data or $30.58$M words.
The phrase length was limited to $5$ and word-alignment was obtained with \textsc{FastAlign} \cite{Dyer+2013:fastalign}.
Lexicalised reordering models were trained using the same data.
For language modelling, we used the complete sets of OOD data (i.e. no data selection).
$5$-gram LMs were trained using \textsc{lmplz} \cite{Heafield+2013:lmplz}.
$100$-best MIRA tuning was employed \cite{Cherry+2012:mira}. For the English-to-French system, tuning was done on the IWSLT 2010 development and evaluation data with a total of 2,551 sentences. For the English-to-German system, tuning was done on the IWSLT 2010 development data with 887 sentences.

In SLT, the input to the MT system was ASR output, which typically lacks casing and punctuation. 
Following previous work \cite{Peitz_iwslt11_puntgen,KIT_iwslt12_segpunctgen}, a monolingual translation system was trained 
to recover casing and punctuation from the ASR output, thus producing source sentences which are 
more adequate for translation. The training data for this monolingual MT system was obtained by pre-processing an actual corpus of the source language to form \textit{pseudo ASR} outputs, which contained no case and punctuation information. Numbers, symbols and acronyms were also converted to their verbal forms with lookup tables. We then used this synthesised corpus of pseudo ASR as the source, and the original corpus as the target of our monolingual MT.
The monolingual translation system was trained on $37.6$M words (Table \ref{table:data_text}). It performed monotonic translation with phrases of as long as 7 words.


\section{Decoding}\label{sec:decoding}

The evaluation systems for ASR and MT are multi-pass systems with resource optimisation and environment management capabilities \cite{Koehn+2007:moses,WEBASR_IS2008}. The ASR is a two-stream multi-pass system. It is illustrated in Figure \ref{fig:ASRsysdiag}. The two streams ASR$_1$ and ASR$_2$ differ by the acoustic model training data (detailed in Table \ref{table:data_am}) and also the tandem configurations (detailed in \S\ref{sec:ASR}). Both streams follow the same routine along the multi-pass decoding system. In pass $1$, a unified decoding result was generated using a non-VTLN DNN and GMM-HMM tandem system with cepstral mean and variance (CMVN) normalisation trained on ASR$_2$ data. These hypothesis transcripts were used for inferring the warp factors. The filterbank (for both ASR$_1$ and ASR$_2$) and PLP (for ASR$_1$ only) features were then warped and CMVN normalised, and the system branched off into two streams with two VTLN decoders trained on ASR$_1$ and ASR$_2$ data respectively.

After pass $2$ decoding, speaker-based MLLR cross adaptations were carried out. The transcripts from ASR$_1$ was used for the model transformation in ASR$_2$ system and vice versa. The number of regression classes was set to $16$. When pass $3$ decoding was done, MLLR self adaptations were performed. The number of regression classes was also set to $16$.

All pre-final stage decoding made use of weighted finite state transducers (WFSTs) for fast implementation. In a pilot experiment, PLP systems with heteroscedastic linear discriminant analysis (HLDA) were trained on the ASR$_2$ data \cite{kumar_specomm98}. WFST decoding with a pruned $4$-gram grammar network was compared with the standard tree search with an unpruned $3$-gram LM. The WER and real-time factor (RT) on IWSLT 2011 evaluation and IWSLT 2012 evaluation data are shown in Table \ref{table:fastdecode}. WFST was shown to achieve the same performance as tree-search decoding, with much faster decoding speed.

\begin{table}
 \caption{Tree-search and WFST decoder\label{table:fastdecode}}
 {\centering \centerline {\resizebox{0.8\linewidth}{!}{
 \begin{tabular}{llrlr}
  \hline
            &  \multicolumn{2}{c}{Tst11} & \multicolumn{2}{c}{Tst12} \\
  Decoder   &  WER  & RT   &  WER  & RT \\
  \hline
  Tree-search   & 23.7\% & 18.4 & 27.0\% & 19.8 \\
  WFST          & 23.7\% & 3.0  & 27.0\% & 3.3 \\
  \hline
 \end{tabular}
 }}}
\end{table}

In the final stage, acoustic and language model rescoring were performed. Base lattices were generated with $2$-gram LM pruned with a threshold $10^{-10}$. Lattice expansion was done with $4$-gram unpruned language models. Three settings were tried and the results were compared,
\begin{compactenum}
 \item[(i)] Language model rescoring with the $4$-gram LM
 \item[(ii)] Considering pronunciation probability (Pron. prob.) on top of (i)
 \item[(iii)] Acoustic and language model rescoring with the setting of (ii)
\end{compactenum}

ASR performance in terms of WER are shown in Table \ref{table:multipassASR}. The initial non-VTLN system gave WER of $16.9\%$ and $17.7\%$ on IWSLT 2011 and 2012 data respectively. Moving towards the VTLN systems, when ASR$_1$ and ASR$_2$ branched off, it is observed that the ASR$_1$ model gave $1.0\%$ to $1.4\%$ lower WER than the ASR$_2$ model. This is because the data in ASR$_1$ had a better match in terms of domain. 
Incremental performance gains can be observed in individual steps, particularly MPE, cross-adaptation and language model rescoring. The WER difference between ASR$_1$ and ASR$_2$ diminished to $0.4$-$0.5\%$ after all optimisation steps. After system combination, the final WER is $21$-$25\%$ relatively lower compared with the initial system.

\begin{table}
 \caption{WER of the multi-pass ASR systems \label{table:multipassASR}}
  {\centering \centerline {\resizebox{0.9\linewidth}{!}{
 \begin{tabular}{lcccc}
  \hline
                 &  \multicolumn{2}{c}{Tst11} & \multicolumn{2}{c}{Tst12} \\
  ASR system     &  ASR$_1$ & ASR$_2$ & ASR$_1$ & ASR$_2$ \\
  \hline
  Non-VTLN        &    --     &  16.9\% &  --       & 17.7\% \\
  $+$VTLN         &   15.4\%  &  16.4\% &  16.4\%   & 16.8\% \\ 
  $+$MPE          &   14.7\%  &  15.7\% &  16.0\%   & 16.1\% \\
  $+$Cross-adapt  &   14.0\%  &  14.9\% &  14.2\%   & 14.8\% \\
  $+$Self-adapt   &   14.0\%  &  15.0\% &  14.2\%   & 14.7\% \\
  $+$LM rescoring &   13.4\%  &  14.5\% &  13.5\%   & 14.2\% \\
  $+$Pron. prob.  &   13.3\%  &  14.2\% &  13.4\%   & 14.0\% \\
  $+$AM rescoring &   13.3\%  &  13.8\% &  13.4\%   & 13.7\% \\
  \addlinespace
  ROVER           & \multicolumn{2}{c}{---13.3\%---} & \multicolumn{2}{c}{---13.2\%---} \\
  \hline
 \end{tabular}
  }}}
  \vspace{-2mm}
\end{table}

\begin{table}[h]
 \caption{MT system performance on eval data \label{table:MTresults}}
 {\centering \centerline {\resizebox{0.92\linewidth}{!}{
 \begin{tabular}{llcc}
  \hline
                &              & \multicolumn{2}{c}{BLEU(c)} \\
  Language pair & \hspace{1mm} & \hspace{2mm}Dev10\hspace{2mm} & \hspace{2mm}Tst12\hspace{2mm} \\
  \hline
  \multicolumn{3}{l}{(MT with true transcript)} \\
  En$\rightarrow$Fr &     &         & $40.9$ \\
  En$\rightarrow$De &     & $21.5$  &        \\
  \addlinespace
  \multicolumn{3}{l}{(Monolingual translation)} \\
  En(pseudo ASR)$\rightarrow$En  &   &  & $88.0$  \\
  En(ASR)$\rightarrow$En         &   &  & $69.0$ \\
  \addlinespace
  \multicolumn{3}{l}{(SLT)} \\
  En(ASR)$\rightarrow$En$\rightarrow$Fr &  &         & $31.7$  \\
  En(ASR)$\rightarrow$En$\rightarrow$De &  & $16.8$  &   \\
  \hline 
 \end{tabular}
 }}}
 \vspace{-2mm}
\end{table}

MT Decoding was performed with cube pruning \cite{Huang+2007:FR} both in tuning and testing. Decoding was done with the minimum Bayes risk criterion and reordering over punctuations was forbidden. To restore the correct case of the output the truecasing heuristic was employed. The same set of standard techniques was applied on En$\rightarrow$Fr and En$\rightarrow$De translation. 

The MT system was tested on IWSLT 2010 development data and 2012 evaluation data, and the results are shown in Table \ref{table:MTresults}. Performance are shown in terms of cased and punctuated BLEU scores. When given the reference transcript, the MT system gave $40.9$ and $21.5$ BLEU score for MT tasks in En$\rightarrow$Fr and En$\rightarrow$De respectively. The monolingual translation system (\S\ref{sec:MT}) restored case and punctuation information. It was tested on pseudo ASR and real ASR output and yielded $88.0$ and $69.0$ BLEU score. Finally in the SLT setting, the decoded ASR result was fed to the monolingual translation system and the output were subsequently translated. The BLEU score is $31.7$ and $16.8$ for SLT tasks in En$\rightarrow$Fr and En$\rightarrow$De respectively. 

In Table \ref{table:SLTresults}, the official IWSLT 2014 evaluation performance in terms of BLEU and TER (cased, punctuated and non-case, non-punctuated) for the USFD primary system is shown.

\begin{table}[b]
 \vspace{-4mm}
 \caption{Primary SLT system performance (Tst14) \label{table:SLTresults}}
 {\centering \centerline {\resizebox{0.99\linewidth}{!}{
 \begin{tabular}{lcccc}
 \hline
  Language pair & \hspace{-1mm}BLEU(c)\hspace{-1mm} & \hspace{-1mm}TER(c)\hspace{-1mm}  & \hspace{-1mm}BLEU\hspace{-1mm}  & \hspace{-1mm}TER\hspace{-1mm} \\
  \hline
  En$\rightarrow$Fr & 23.45 & 59.94 & 24.14 & 58.97 \\
  En$\rightarrow$De & 14.75 & 70.15 & 15.24 & 69.15 \\
  \hline
 \end{tabular}
 }}}
 \vspace{-3mm}
\end{table}


\begin{figure}[h]
 \includegraphics[width=0.95\linewidth]{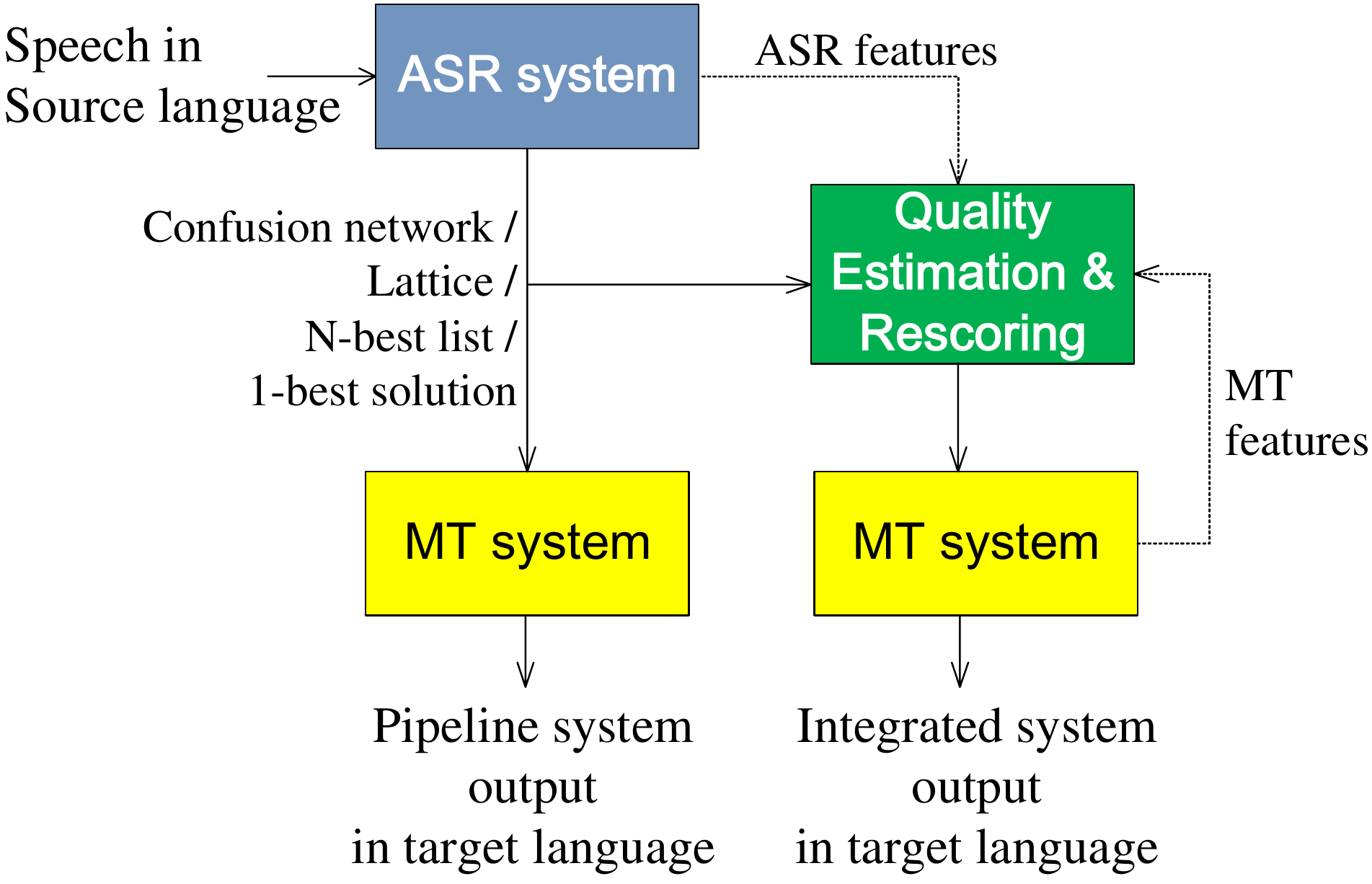}
 \caption{System integration with ASR and MT\label{fig:integration}}
\end{figure}

\section{System integration}\label{sec:sysintegration}

The USFD primary system is a pipeline SLT system in which 1-best ASR result was directly fed to the MT system. System integration experiments were tried in the En$\rightarrow$Fr SLT task and the results were submitted as contrastive systems. Figure \ref{fig:integration} depicts the integrated system and its comparison with the pipeline system. In the integrated system, ASR system hypotheses are expanded in the form of lattices, confusion networks or $N$-best lists. 
A quality estimation (QE) module evaluated and rescored the ASR outputs before they were fed to the MT system. 

In our implementation, $10$-best outputs from the ASR system on the IWSLT 2011 evaluation data were used for QE training. The QE module derived $117$ QuEst \cite{Specia2013,ShahABiciciS13} features from each sentence to describe its linguistic, statistical properties as well as the statistics from the ASR and MT models. Out of the $117$ features, top $58$ features were selected using the Gaussian Process (GP) with RBF kernel as described in \cite{Shah2013}. Further, GP was used to learn the relationship between the selected features and the translation performance of the sentence (in this case, sentence-based METEOR score) \cite{denkowski:lavie:meteor-wmt:2014}. During testing, the estimated translation performance was used to rescore the $10$-best ASR output. Details of the integrated system were described in \cite{Ng_2015ICASSP_QE}.

\begin{table}[h]
\caption{Contrastive SLT system performance (En$\rightarrow$Fr)\label{table:QE}}
{\centering \centerline{\resizebox{0.99\linewidth}{!}{
 \begin{tabular}{lcc}
 \hline
 Setting  & \hspace{0mm}Tst12\hspace{0mm}   & \hspace{0mm}Tst14\hspace{0mm} \\
 \hline
 Contrastive 1  {\footnotesize (baseline)}                            & $31.33$  & $23.18$ \\
 Contrastive 2  \\
 {\footnotesize ($+$ $10$-best list rescoring)}                       & $31.51$  & $23.27$ \\
 Contrastive 3 \\
 {\footnotesize ($+$ ASR confidence-informed rescoring)}\hspace{-2mm} & $31.87$  & $23.44$ \\
 \hline
 \end{tabular}
}}}
\end{table}

The ROVER combination of ASR$_1$ and ASR$_2$ systems only provided $1$-best output. In the integration experiment, the $10$-best output from ASR$_1$ was used instead. 

Performance of the contrastive systems in terms of cased and punctuated BLEU score is shown in Table \ref{table:QE}. 
Contrastive 1 result is from the baseline system with pipeline setting. Contrastive 2 and 3 show the results of two different system integration settings. The baseline system gave BLEU scores $31.33$ and $23.18$ on IWSLT 2012 and IWSLT 2014 data. The baseline numbers are inferior to the primary system number (IWSLT 2012: $31.7$; IWSLT 2014: $23.45$) as shown in Table \ref{table:MTresults} and \ref{table:SLTresults}. This is because the baseline here did not benefit from ASR system combination.

Rescoring gives $0.18$ and $0.09$ BLEU improvements to IWSLT 2012 and IWSLT 2014 data respectively.
By inspecting the results, it was found that rescoring generally had higher effectiveness for the sentences with low ASR confidence. Therefore, a confidence threshold was set, and rescoring was only performed when the ASR confidence dropped below this threshold. For IWSLT 2012 data, optimality was reached when 55\% of the sentences were selected by this confidence criteria to rescore, resulting a further $0.36$ BLEU score gain. This threshold was applied on IWSLT 2014 data, a $0.17$ BLEU score gain was observed.

\section{Summary}

In this paper, the USFD SLT system for IWSLT 2014 was described. Automatic speech recognition (ASR) is achieved by two multi-pass deep neural network systems with slightly different tandem configurations and different training data. Machine translation (MT) is achieved by a monolingual phrase-based monotonic translation system which recovers case and inserts punctuation, followed by a bilingual phrase-based translation system. The USFD contrastive systems explore the integration of ASR and MT by using a quality estimation system to rescore the ASR outputs, optimising towards better translation. This gives  noticeable BLEU improvement on the IWSLT 2012 and 2014 evaluation data.


%
\bibliographystyle{IEEEtran}
\bibliography{USFDiwslt2014}

\end{document}